\documentclass[10pt,twocolumn,letterpaper]{article}

\usepackage{mmstyle}
\usepackage{cvpr}
\usepackage{times}
\usepackage{epsfig}
\usepackage{graphicx}
\usepackage{amsmath}
\usepackage{amssymb}
\usepackage{algorithm}
\usepackage{algorithmicx}
\usepackage{algpseudocode}
\usepackage{makecell}
\usepackage{textcomp}
\usepackage{subcaption}

\newcommand{\misscite}{\textcolor{red}{[C]~}}

\usepackage[pagebackref=true,breaklinks=true,letterpaper=true,colorlinks,bookmarks=false]{hyperref}

\cvprfinalcopy % *** Uncomment this line for the final submission

\def\cvprPaperID{1510} % *** Enter the CVPR Paper ID here

\ifcvprfinal\fi
\begin{document}

%%%%%%%%% TITLE
\title{Learning to Cluster Faces on an Affinity Graph}

\author{Lei Yang,\textsuperscript{1}
Xiaohang Zhan,\textsuperscript{1}
Dapeng Chen,\textsuperscript{2}
Junjie Yan,\textsuperscript{2}
Chen Change Loy,\textsuperscript{3}
Dahua Lin,\textsuperscript{1} \\
\textsuperscript{1}CUHK - SenseTime Joint Lab, The Chinese University of Hong Kong \\
\textsuperscript{2}SenseTime Group Limited,
\textsuperscript{3}Nanyang Technological University \\
{\tt\small \{yl016, zx017, dhlin\}@ie.cuhk.edu.hk, \{chendapeng, yanjunjie\}@sensetime.com, ccloy@ntu.edu.sg}
%\and
}

\maketitle
\thispagestyle{empty}

%%%%%%%%% ABSTRACT
% !TEX root = ../submission.tex

\begin{abstract}
Face recognition sees remarkable progress in recent years, and its
performance has reached a very high level.
Taking it to a next level requires substantially larger data, which
would involve prohibitive annotation cost.
Hence, exploiting unlabeled data becomes an appealing alternative.
Recent works have shown that clustering unlabeled faces is a promising approach,
often leading to notable performance gains.
Yet, how to effectively cluster, especially on a large-scale (\ie~million-level
or above) dataset, remains an open question. A key challenge lies in the complex
variations of cluster patterns, which make it difficult for conventional
clustering methods to meet the needed accuracy.
This work explores a novel approach, namely, learning to cluster instead
of relying on hand-crafted criteria. Specifically, we propose a framework
based on graph convolutional network, which combines a detection and
a segmentation module to pinpoint face clusters.
Experiments show that our method yields significantly more accurate face
clusters, which, as a result, also lead to further performance gain in
face recognition.
\end{abstract}

%%%%%%%%% BODY TEXT
% !TEX root = ../submission.tex

\section{Introduction}

%% Background: large data set, high annotation cost

Thanks to the advances in deep learning techniques,
the performance of face recognition has been remarkably boosted~\cite{sun2014deep,schroff2015facenet,wang2018cosface,deng2018arcface,zhang2018accelerated}.
However, it should be noted that the high accuracy of modern face recognition
systems relies heavily on the availability of large-scale annotated training
data. While one can easily collect a vast quantity of facial images from
the Internet, annotating them is prohibitively expensive.
Therefore, exploiting unlabeled data, \eg~through unsupervised or
semi-supervised learning, becomes a compelling option and has attracted
lots of interest from both academia and industry~\cite{zhan2018consensus,otto2018clustering}.

% Though face recognition has achieved great progress with high-capacity models
% and large-scale labeled data~\misscite, there is still a long way to go for
% the current industrial requirements like searching a person in tens of millions
% images. To achieve better performance, it requires substantially larger-scale
% data. Existing large-scale face recognition datasets including MS-Celebrity-1M
% and MegaFace make training on million-level data possible. However, they suffer
% from severe annotation noises~\misscite and data cleaning is always laborious.
% On the other hand, massive unlabeled face images are easy to collect, either by
% web crawler or through security surveillance.

\begin{figure}[t]
	\centering
	\includegraphics[width=0.9\linewidth]{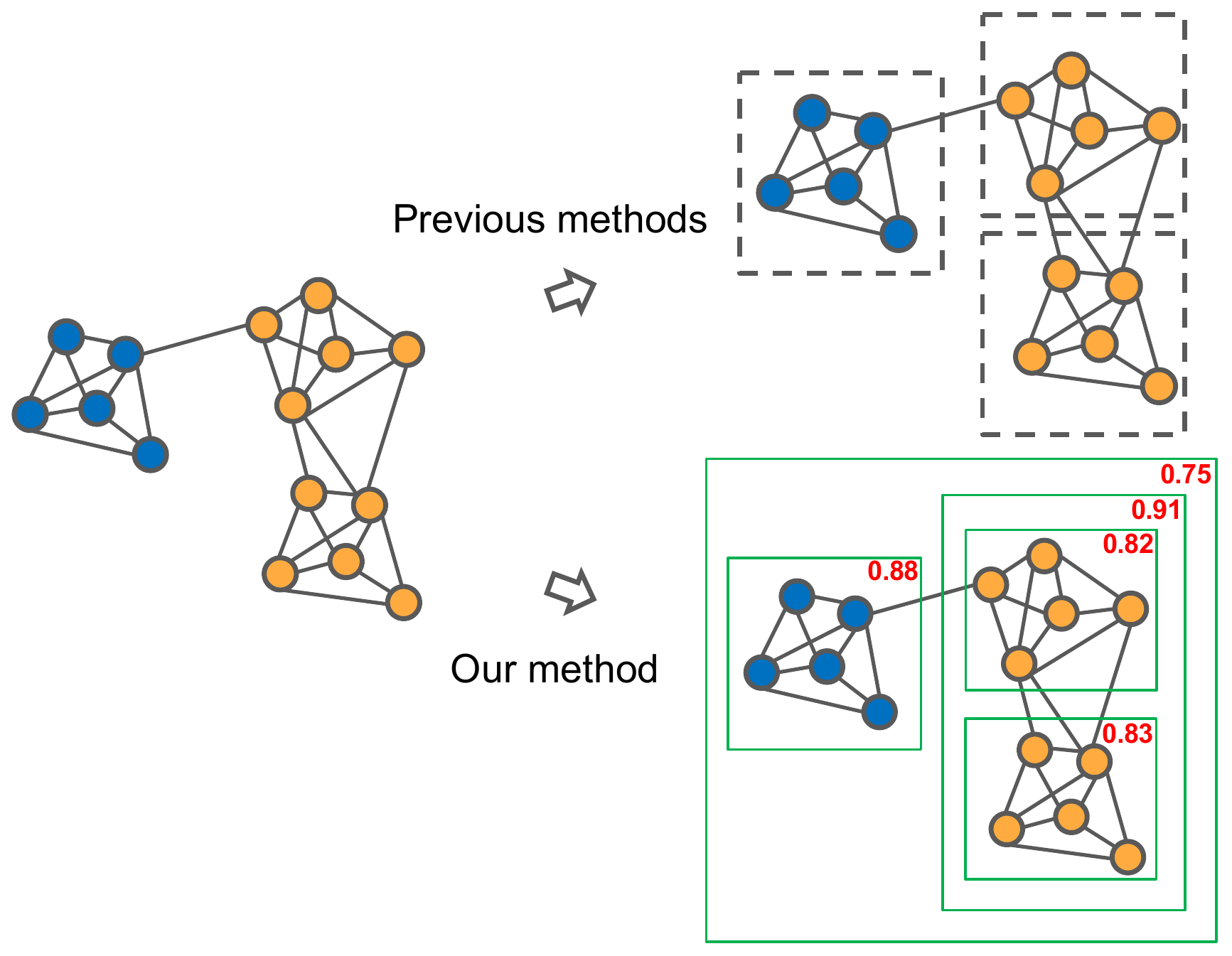}
	\caption{\small
        A case to demonstrate the difference between existing methods
        and our approach. The blue vertices and orange vertices represent
        two classes respectively. Previous unsupervised methods relying on
        specific clustering policies may not be able to handle the orange cluster with
        complex intra-structures. While our approach, through learning
        from the structures, is able to evaluate different combinations of
        cluster proposals (green boxes) and output the clusters with high scores.
    }
	\label{fig:case}
\end{figure}

%% Unsupervised clustering -- limitations

A natural idea to exploit unlabeled data is to cluster them into
``pseudo classes'', such that they can be used like labeled data and fed to
a supervised learning pipeline. Recent works~\cite{zhan2018consensus} have shown that
this approach can bring performance gains.
Yet, current implementations of this approach still leave a lot to be desired.
Particularly, they often resort to unsupervised methods,
such as K-means~\cite{lloyd1982least},
spectral clustering~\cite{ho2003clustering},
hierarchical clustering~\cite{zhao2006automatic},
and approximate rank-order~\cite{otto2018clustering},
to group unlabeled faces.
These methods rely on simplistic assumptions,
\eg, K-means implicitly assumes that the samples in each cluster are
around a single center; spectral clustering requires  that the
cluster sizes are relatively balanced, etc.
Consequently, they lack the capability of coping with complicated
cluster structures, thus often giving rise to noisy clusters, especially
when applied to large-scale datasets collected from real-world settings.
This problem seriously limits the performance improvement.

Hence, to effectively exploit unlabeled face data, we need to develop an
effective clustering algorithm that is able to cope with the complicated
cluster structures arising frequently in practice. Clearly, relying on
simple assumptions would not provide this capability.
In this work, we explore a fundamentally different approach, that is,
to \emph{learn} how to cluster from data.
Particularly, we desire to draw on the strong expressive power of graph
convolutional network to capture the common patterns in face clusters,
and leverage them to help to partition the unlabeled data.

%% Approach overview

We propose a framework for face clustering based on
graph convolutional networks~\cite{kipf2017semi}.
This framework adopts a pipeline similar to the Mask R-CNN~\cite{he2017mask}
for instance segmentation,
\ie, generating proposals, identifying the positive ones, and then refining
them with masks.
These steps are accomplished respectively by an iterative
proposal generator based on super-vertex, a graph detection network,
and a graph segmentation network.
It should be noted that while we are inspired by Mask R-CNN,
our framework still differs essentially: the former operates on a 2D image grid
while the latter operates on an affinity graph with arbitrary structures.
As shown in Figure~\ref{fig:case}, relying on the structural patterns learned
based on a graph convolutional network instead of some simplistic assumptions,
our framework is able to handle clusters with complicated structures.

% Instead of designing an unsupervised method, we turn to learn the characteristic
% of a desired cluster based on the affinity graph and the Graph Convolutional
% Networks~\cite{kipf2017semi}. The affinity graph of unlabeled data contains not
% only unary information but important pairwise relationships, while the GCNs show
% great potential to learn from structural data~\misscite{} from the affinity
% graph. To effectively pinpoint clusters from a set of cluster proposals, we
% propose a GCN based detection and segmentation framework, which is inspired by
% object detection and instance segmentation task in images~\misscite{}, The
% framework mainly contains two modules, the graph detection net and the graph
% segmentation net, which aims at discovering rough clusters and removing noise
% nodes, respectively. The two stage paradigm is similar to instance segmentation
% using Mask R-CNN~\cite{he2017mask}. As shown in Figure~\ref{fig:case}, different
% from previous methods that create clusters with a specific policy which may not
% be able to handle clusters with complex intra-structures, our approach can learn
% from the structures with GCNs to evaluate different combinations of cluster
% proposals and output the clusters with high scores.

%% Experimental results

The proposed method significantly improves the clustering accuracy on
large-scale face data, achieving a F-score at $85.66$, which is not only
superior to the best result obtained by unsupervised clustering methods
(F-score $68.39$) but also higher than a recent
state of the art~\cite{zhan2018consensus} (F-score $75.01$).
Using this clustering framework to process the unlabeled data,
we improve the performance of a face recognition model on MegaFace from $60.29$ to
$78.64$, which is quite close to the performance obtained by
supervised learning on all the data ($80.75$).

%  clustering performance with F-score
% \missvalue, comparing to the baseline~\cite{zhan2018consensus} which
% achieves F-score \missvalue. The performance is also much better than the
% representative unsupervised method yielding f-score \missvalue ~\misscite{}.
% Afterwards, we assign new labels for discovered clusters and utilize them to
% update the base face recognition model. We improve the performance of the base
% model on MegaFace benchmark from \missvalue to \missvalue, the performance is
% very close to the supervised upper bound when all the data are labeled.

%% Contributions

The main contributions lie in three aspects:
(1) We make the first attempt to perform top-down face clustering in a
supervised manner.
(2) It is the first work that formulates clustering as a detection
and segmentation pipeline based on graph convolution networks.
(3) Our method achieves state-of-the-art performance in large-scale
face clustering, and boosts the face recognition model close to the
supervised result when applying the discovered clusters.
% !TEX root = ../submission.tex

\section{Related Work}

\paragraph{Face Clustering}
Clustering is a basic task in machine learning.
Jain~\etal~\cite{jain2010data} provide a survey for classical clustering
methods. Most existing clustering methods are unsupervised.
Face clustering provides a way to exploit massive unlabeled data.
The study along this direction remains at an early stage. The question of
how to cluster faces on large-scale data remains open.

Early works use hand-crafted features and classical clustering algorithms.
For example,
Ho~\etal~\cite{ho2003clustering} used gradient and pixel intensity
as face features.
Cui~\etal~\cite{cui2007easyalbum} used LBP features.
Both of them adopt spectral clustering.
Recent methods make use of learned features.
\cite{kaufman2009finding} performed top-down clustering in an unsupervised way.
Finley~\etal~\cite{finley2005supervised} proposed an SVM-based
supervised method in a bottom-up manner.
Otto~\etal~\cite{otto2018clustering} used deep features from a CNN-based
face model and proposed an approximate rank-order metric to link
images pairs to be clusters.
Lin~\etal~\cite{lin2017proximity} designed a similarity measure based
on linear SVM trained on the nearest neighbours of data samples.
Shi~\etal~\cite{shi2018face} proposed Conditional Pairwise Clustering,
formulating clustering as a conditional random field to
cluster faces by pair-wise similarities.
Lin~\etal~\cite{lin2018deep} proposed to %Deep Density Clustering, which
exploit local structures of deep features by introducing minimal
covering spheres of neighbourhoods to improve similarity measure.
Zhan~\etal~\cite{zhan2018consensus} trained a MLP classifier to aggregate
information and thus discover more robust linkages,
then obtained clusters by finding connected components.

Though using deep features, these works mainly concentrate on designing
new similarity metrics, and still rely on unsupervised methods to
perform clustering.
Unlike all the works above, our method \emph{learns} how to cluster in a top-down manner,
based on a detection-segmentation paradigm. This allows the model to
handle clusters with complicated structures.

\vspace{-11pt}
\paragraph{Graph Convolutional Networks}
Graph Convolutional Networks (GCNs)~\cite{kipf2017semi} extend
CNNs to process graph-structured data.
Existing work has shown the advantages of GCNs, such as the strong
capability of modeling complex graphical patterns. On various tasks,
the use of GCNs has led to considerable performance
improvement~\cite{kipf2017semi, hamilton2017inductive, van2017graph, yan2018spatial}.
%
%It is better than the neural networks that only makes use of the
%	sample features，and it is also superior to the graph embedding
%	approaches that do not use the features nodes
%
For example,
Kipf~\etal~\cite{kipf2017semi} applied the GCNs to semi-supervised classification.
Hamilton~\etal~\cite{hamilton2017inductive} leveraged GCNs to learn feature
representations.
Berg~\etal~\cite{van2017graph} showed that GCNs are superior to other methods
in link prediction.
Yan~\etal~\cite{yan2018spatial} employed GCNs to model human joints for
skeleton-based action recognition.

%
%In this paper, we follow the spirit of GCNs and adapt it to both
%	evaluate the cluster quality and analyze the noise inside a cluster.

In this paper, we adopt GCN as the basic machinery to capture cluster
patterns on an affinity graph. To our best knowledge, this is the first
work that uses GCN to learn how to cluster in a supervised way.

% !TEX root = ../submission.tex

\section{Methodology}

\begin{figure*}[t]
	\centering
	\includegraphics[width=0.72\linewidth]{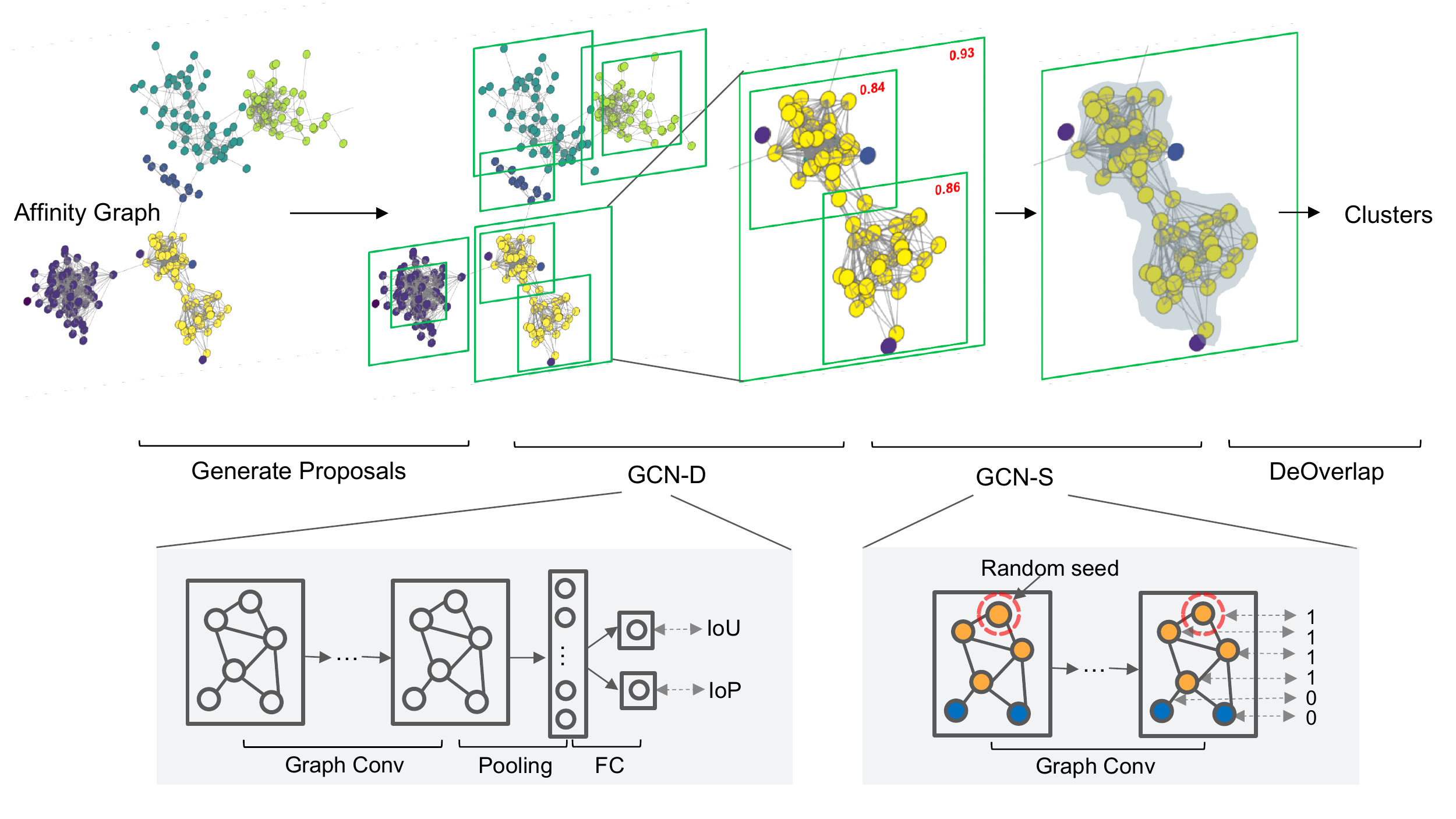}
	\caption{\small
        Overview of graph detection and segmentation for clustering.
    }
	\label{fig:pipeline}
\end{figure*}

In large-scale face clustering, the complex variations of the cluster
patterns become the main challenge for further performance gain.
To tackle the challenge, we explore a supervised approach, that is, to
learn the cluster patterns based on graph convolutional networks.
Specifically, we formulate this as a joint detection and segmentation
problem on an affinity graph.

Given a face dataset, we extract the feature for each face image
with a trained CNN, forming a set of features
$\cD=\{\vf_i\}_{i=1}^{N}$, where $\vf_{i}$ is a $d$-dimensional vector.
To construct the affinity graph, we regard each sample as a vertex
and use cosine similarity to find $K$ nearest neighbors for each sample.
By connecting between neighbors,
we obtain an affinity graph $\cG=(\cV, \cE)$ for the whole dataset.
Alternatively, the affinity graph $\cG$ can also be represented by
a symmetric adjacent matrix $\mA\in \mathbb{R}^{N \times N}$,
where the element $a_{i,j}$ is the cosine similarity between
$\vf_{i}$ and $\vf_{j}$ if two vertices are connected, or zero
otherwise.
The affinity graph is a large-scale graph with millions of vertices.
From such a graph, we desire to find clusters that have the following
properties:
(1) different clusters contain the images with different labels; and
(2) images in one cluster are with the same label.

% \if 0
% To build the affinity graph, we extract the features with a trained
% 	CNN for all the images in the dataset,
% 	forming $\cD=\{f_{i}\}_{i=1}^{N}$. where $f_{i}$ is a
% 	$d$-dimensional vector.
% %
% For every feature in $\cD$, we find its $K$ nearest neighbors by
% 	cosine similarity, denoted by $\cN_{i}$ for feature $f_{i}$.
% %
% The affinity graph $\cG=(\cV, \cE)$ for the whole dataset is
% 	constructed according to the neighbours of all the features in $\cD$.
% %
% More specifically, $\cG$ is a undirected graph. Its nodes
% 	$\cV$ are defined over all the features, \emph{i.e.},
% 	$\cV=\{v_{i}\}_{i=1}^{N}$, while its edges $\cE$ connect every
% 	feature and its neighbours, \emph{i.e.},
% 	$\cE=\{ e_{i,j}| f_{j} \in \cN_{i} \}_{i=1}^{N}$. The weight for
% 	each edge $e_{i, j}$ is computed by the cosine similarity between
% 	$f_{i}$ and $f_{j}$, denoted by $w_{i,j} = \langle f_{i}, f_{j} \rangle$.
% %
% With the edge weights, the affinity graph $\cG$ can also be
% 	represented by a symmetric adjacent matrix
% 	$\mA\in \mathbb{R}^{N \times N}$, where the element
% 	$a_{i,j} = w_{i,j}$ if $v_{i}$ and $v_{j}$ are connected,
% 	otherwise $a_{i,j} = 0$.
% \fi

\subsection{Framework Overview}

As shown in Figure~\ref{fig:pipeline}, our clustering framework
consists of three modules, namely proposal generator, GCN-D, and GCN-S.
The first module generates cluster proposals,
\ie, sub-graphs likely to be clusters, from the affinity graph.
With all the cluster proposals, we then introduce two GCN modules,
GCN-D and GCN-S, to form a two-stage procedure,
which first selects high-quality proposals and then refines the selected
proposals by removing the noises therein.
Specifically,
GCN-D performs cluster detection.
Taking a cluster proposal as input, it evaluates
how likely the proposal constitutes a desired cluster.
Then GCN-S performs the segmentation to refine the selected proposals.
Particularly, given a cluster, it estimates the probability of being noise
for each vertex, and prunes the cluster by discarding the outliers.
According to the outputs of these two GCNs, we can efficiently obtain
high-quality clusters.

\subsection{Cluster Proposals}

Instead of processing the large affinity graph directly,
we first generate cluster proposals.
It is inspired by the way of generating region proposals
in object detection~\cite{girshick2014rich, girshick2015fast}.
Such a strategy can substantially reduce the computational cost,
since in this way, only a limited number of cluster candidates need to be
evaluated.
A cluster proposal $\cP_{i}$ is a sub-graph of the affinity graph
$\cG$. All the proposals compose a set $\cP=\{\cP_{i}\}_{i=1}^{N_{p}}$.
The cluster proposals are generated based on super-vertices,
and all the super-vertices form a set $\cS=\{\cS_{i}\}_{i=1}^{N_{s}}$.
In this section, we first introduce the generation of super-vertex,
and then devise an algorithm to compose cluster proposals thereon.

% \if 0
% \noindent \textbf{Super-Vertex.} Compared with the desired clusters,
% 	the super-vertex is a kind of more conservative presentation.
% 	Images in a super-vertex are expected to describe a same person,
% 	but different super-vertices may still describe a same person.
% %
% Intuitively, images in a super-vertex are similar to each other, thus
% 	they are closely connected in the affinity graph $\cG$. It is
% 	straightforward to use \emph{connected component} in graph
% 	theory to represent super-vertex.
% %
% However, connected component on the affinity graph is likely to be
% 	too large to keep the most prominent information.
% %
% To control the intra-consistency of super-vertex, we introduce
% 	a edge threshold $e_\tau$ on the affinity graph and maximum size
% 	$s_{max}$ of super-vertex. Alg.~\ref{alg:super_vertex} shows the
% 	detailed procedure to produce the super-vertex set $\cS$ .
% \fi

\noindent \textbf{Super-Vertex.}
A super-vertex is a sub-graph containing a small number of vertices
that are closely connected to each other.
Hence, it is natural to use connected components
to represent super-vertex.
However, the connected component directly derived from the graph $\cG$
can be overly large.
To maintain high connectivity within each super-vertice, we remove those
edges whose affinity values are below a threshold $e_\tau$ and constrain
the size of super-vertices below a maximum $s_{max}$.
Alg.~\ref{alg:super_vertex} shows the detailed procedure to produce the
super-vertex set $\cS$.
Generally, an affinity graph with $1M$ vertices can be partitioned into $50K$
super-vertices, with each containing $20$ vertices on average.

\noindent \textbf{Proposal Generation.}
Compared with the desired clusters, the super-vertex is a conservative
formation. Although the vertices in a super-vertex are highly possible to describe
the same person, the samples of a person may distribute into several super-vertices.
Inspired by the multi-scale proposals in object detection~\cite{girshick2014rich,girshick2015fast},
we design an algorithm to generate multi-scale cluster proposals.
As Alg.~\ref{alg:proposal} shows, we construct a higher-level graph on top of
the super-vertices, with the centers of super-vertices as the vertices and
the affinities between these centers as the edges.
With this higher-level graph, we can apply Alg.~\ref{alg:super_vertex} again
and obtain proposals of larger sizes.
By iteratively applying this construction for $I$ times,
we obtain proposals with multiple scales.
%
%The impact of the cluster proposal strategy will be investigated in the
%ablation study (sec.~\ref{sec:abl_props}).

\begin{algorithm}[t]
 \caption{Super-vertex Generation}
 \begin{algorithmic}[1]
 \renewcommand{\algorithmicrequire}{\textbf{Input:}}
 \renewcommand{\algorithmicensure}{\textbf{Output:}}
 \Require Affinity Graph $\mA$, edge threshold $e_\tau$, maximum size $s_{max}$, threshold step $\Delta$.
 \Ensure Super-Vertices $\cS$

   \State $\cS = \emptyset, \cR = \emptyset$
  \State $\cC, \cR$ = \Call{FindSuperVertices}{$\mA$, $e_\tau$, $s_{max}$}
  \State $\cS = \cS \cup \cC$
  \While{$\cR \neq \emptyset$}
   \State $e_\tau = e_\tau + \Delta$
   \State $\cC, \cR$ = \Call{FindSuperVertices}{$\mA_{\cR}$, $e_\tau$, $s_{max}$}
   \State $\cS = \cS \cup \cC$
  \EndWhile
  \State \Return $\cS$
    \Statex
    \Function{FindSuperVertices}{$\mA$, $e_\tau$, $s_{max}$}
   \State $\mA'$ = \Call{PruneEdge}{$\mA$, $e_\tau$}
   \State $\cX$ = \Call{FindConnectedComponents}{$\mA'$}
   \State $\cC = \{c | c \in \cX, |c| < s_{max} \}$
   \State $\cR = \cX \setminus \cC$
   \State \Return $\cC, \cR$
  \EndFunction
 \end{algorithmic}
 \label{alg:super_vertex}
\end{algorithm}

\begin{algorithm}[t]
 \caption{Iterative Proposal Generation}
 \renewcommand{\algorithmicrequire}{\textbf{Input:}}
 \renewcommand{\algorithmicensure}{\textbf{Output:}}
 \begin{algorithmic}[1]
 \Require Super-Vertex set $\cS$, Iterative Number $I$, edge threshold $e_\tau$, maximum size $s_{max}$, threshold step $\Delta$.
 \Ensure Proposal set $\cP$
  \State $\cP = \emptyset, i = 0, \cS'=\cS$
  \While{$i < I$}
   \State $\cP = \cP \cup \cS'$
   \State \parbox[t]{0.9\linewidth}{$\cD = \{f_{s} | s \in \cS' \}$, where $f_{s}$ is the average feature of the vertices in $s$.}
   \vspace{0.3em}
   \State $\mA$ = \Call{BuildAffinityGraph}{$\cD$}
   \State $\cS'$ = \Call{Algorithm 1}{$\mA, e_\tau, s_{max}, \Delta$}
   \State $i = i + 1$
  \EndWhile
  \State \Return $\cP$
 \end{algorithmic}
 \label{alg:proposal}
\end{algorithm}

\subsection{Cluster Detection}

We devise \emph{GCN-D}, a module based on a graph convolutional network (GCN),
to select high-quality clusters from the generated cluster proposals.
Here, the quality is measured by two metrics, namely \emph{IoU} and \emph{IoP}
scores. Given a cluster proposal $\cP$, these scores are defined as
\begin{equation}
IoU(\cP) = \dfrac{|\cP \cap \widehat{\cP}|}{|\cP \cup \widehat{\cP}|}, \quad
IoP(\cP) = \dfrac{|\cP \cap \widehat{\cP}|}{|\cP|},
\label{eq:iou}
\end{equation}
where $\widehat{\cP}$ is the ground-truth set comprised all the vertices
with label $l(\cP)$, and $l(\cP)$ is the \emph{majority label} of the cluster
$\cP$, \ie~the label that occurs the most in $\cP$.
Intuitively, \emph{IoU} reflects how close $\cP$ is to the desired
ground-truth $\widehat{\cP}$; while \emph{IoP} reflects the purity,
\ie~the proportion of vertices in $\cP$ that are with the majority label
$l(\cP)$.

\vspace{-12pt}
\paragraph{Design of GCN-D.}
We assume that high quality clusters usually exhibit certain structural patterns
among the vertices.
We introduce a GCN to identify such clusters. Specifically, given a cluster
proposal $\cP_i$, the GCN takes the visual features associated with its
vertices (denoted as $\mF_0(\cP_i)$) and the affinity sub-matrix (denoted as
$\mA(\cP_i)$) as input, and predicts both the IoU and IoP scores.

 The GCN networks consist of $L$ layers and
 the computation of each layer can be formulated as:
 \begin{equation}
	\mF_{l+1}(\cP_{i}) = \sigma\left(
		\tilde{\mD}(\cP_{i})^{-1}(\mA(\cP_i) + \mI) \mF_l(\cP_{i})\mW_l
	\right),
	\label{eq:gcn}
 \end{equation}
where
$\tilde{\mD} = \sum_j \tilde{\mA}_{ij}(\cP_{i})$ is a diagonal degree matrix.
$\mF_l(\cP_{i})$ contains the embeddings of the $l$-th layer.
$\bold{W}_{l}$ is a matrix to transform the embeddings and
$\sigma$ is the nonlinear activation function (\emph{ReLU} is chosen in this work).
Intuitively, this formula expresses a procedure of taking weighted average
of the embedded features of each vertex and its neighbors, transforming them
with $\bold{W}_{l}$, and then feeding them through a nonlinear activation.
This is similar to a typical block in CNN, except that it operates on a graph
with arbitrary topology.
On the top-level embeddings $\mF_{L}(\cP_{i})$,
we apply a max pooling over all the vertices in $\cP_{i}$,
and obtain a feature vector that provides an overall summary.
Two fully-connected layers are then employed to predict the IoU and
IoP scores, respectively.

% The aggregation processes of GCNs leverage local neighbourhood
% 	information, \ie local graph structure and vertex feature, when
% 	constructing the hidden embedding for each node.
% %
% Note that isomorphic graphs will be
% 	mapped to the same representation and output the same prediction.

\vspace{-12pt}
\paragraph{Training and Inference.}
Given a training set with class labels,
we can obtain the ground-truth IoU and IoP scores
following Eq.\eqref{eq:iou} for each cluster proposal $\cP_{i}$.
Then we train the GCN-D module, with the objective to minimize the
\emph{mean square error(MSE)} between ground-truth and predicted
scores.
We experimentally show that, without any fancy techniques,
GCN can give accurate prediction.
During inference, we use the trained GCN-D to predict both the
IoU and IoP scores for each proposal.
The IoU scores will be used in sec.~\ref{sec:deoverlap} to first retain
proposals with high IoU.
%Then, we rank the proposals by IoU, retaining a fixed number
%of top proposals.
The IoP scores will be used in the next stage
to determine whether a proposal needs to be refined.

\subsection{Cluster Segmentation}

\begin{figure}[t]
	\centering
	\includegraphics[width=0.9\linewidth]{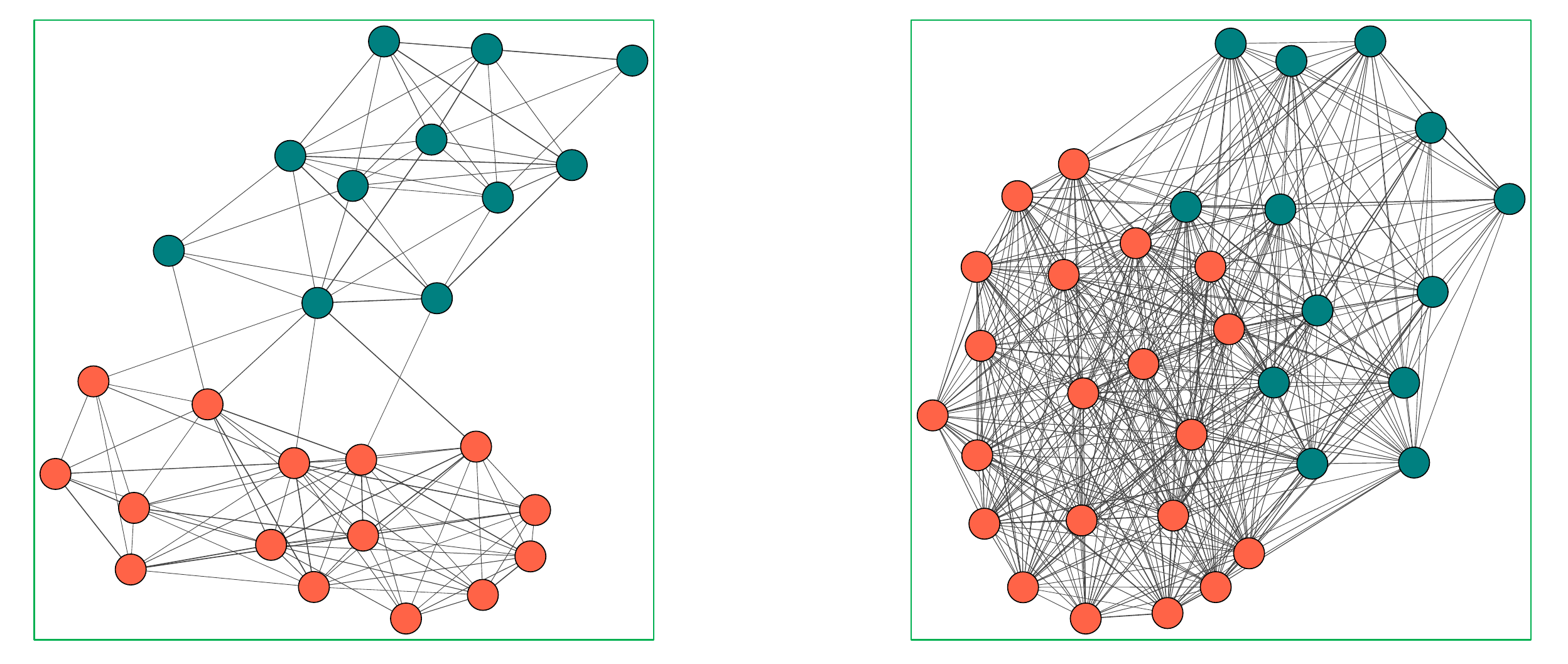}
	\caption{\small
        This figure shows two noisy proposals. Vertices of the same
        color belong to the same ground-truth class. The
        number of red vertices is slightly larger than the number of
        green vertices. As outliers are defined with respect to the
        proposal, \ie, not absolute outliers in a proposal,
        both red and green segmentation results are correct.
        (Best viewed in color.)
    }
	\label{fig:demo_gcn_s}
\end{figure}

The top proposals identified by GCN-D may not be completely pure.
These proposals may still contain a few \emph{outliers},
which need to be eliminated.
To this end, we develop a cluster segmentation module, named
\emph{GCN-S}, to exclude the outliers from the proposal.

\vspace{-11pt}
\paragraph{Design of GCN-S.}
The structure of \emph{GCN-S} is similar to that of \emph{GCN-D}.
The differences mainly lie in the values to be predicted.
Instead of predicting quality scores of an entire cluster $\cP$,
GCN-S outputs a probability value for each vertex $v$ to indicate how
likely it is a genuine member instead of an outlier.

\vspace{-11pt}
\paragraph{Identifying Outliers}
To train the GCN-S, we need to prepare the ground-truth, \ie~
identifying the outliers. % This is nontrivial.
A natural way is to treat all the vertices whose labels are different
from the majority label as outliers.
However, as shown in Fig.~\ref{fig:demo_gcn_s}, this way may encounter
difficulties for a proposal that contains an almost equal number of vertices
with different labels.
To avoid overfitting to manually defined outliers,
we encourage the model to learn different segmentation patterns.
As long as the segmentation result contains vertices from one class,
no matter it is majority label or not, it is regarded as a reasonable solution.
Specifically, we randomly select a vertex in the proposal as
the seed. We concat a value to each vertex feature, where the value of
the selected seed is one while others is zero.
The vertices that have the same label with the seed are
regarded as the positive vertices while others are considered as outliers.
We apply this scheme multiple times with randomly chosen seeds and
thus acquire multiple training samples from each proposal $\cP$.
%that may be annotated differently.

\vspace{-11pt}
\paragraph{Training and Inference.}
With the process above, we can prepare a set of training samples
from the retained proposals. Each sample contains
a set of feature vectors, each for a vertex,
an affinity matrix, as well as a binary vector to indicate whether
the vertices are positive or not.
Then we train the GCN-S module, using the
vertex-wise binary cross-entropy as the loss function.
During inference, we also draw multiple hypotheses for a generated cluster
proposal, and only keep the predicted results that have the most positive
vertices (with a threshold of $0.5$).
This strategy avoids being misled by the case where a vertex
associated with very few positive counterparts is chosen
as the seed.

We only feed the proposals with \emph{IoP} between $0.3$ and $0.7$
to GCN-S.
Because when the proposal is very pure, the outliers are usually
hard examples that need not be removed.
When the proposal is very impure, it is probable that none of the classes
dominate, therefore the proposal might not be suitable to be processed by GCN-S.
With the GCN-S predictions, we remove the outliers from the proposals.

\subsection{De-Overlapping}
\label{sec:deoverlap}

The three stages described above result in a collection of clusters.
However, it is still possible that different clusters may overlap,
\ie~sharing certain vertices. This may cause an adverse effect to
the face recognition training performed thereon.
Here, we propose a simple and fast \emph{de-overlapping} algorithm
to tackle this problem.
Specifically, we first rank the cluster proposals in
descending order of IoU scores. We sequentially collect the proposals
from the ranked list, and modify each proposal by removing the
vertices seen in preceding ones.
The detailed algorithm is described in Alg.~\ref{alg:de_overlap}.

Compared to the Non-Maximum Suppression (NMS) in object detection,
the de-overlapping method is more efficient. Particularly,
the former has a complexity of $O(N^2)$, while the latter has
$O(N)$.
This process can be further accelerated by setting a threshold of
\emph{IoU} for de-overlapping.

\begin{algorithm}[t]
 \caption{De-overlapping}
 \begin{algorithmic}[1]
 \renewcommand{\algorithmicrequire}{\textbf{Input:}}
 \renewcommand{\algorithmicensure}{\textbf{Output:}}
 \Require Ranked Cluster Proposals $\{ \cP'_{0}, \cP'_{1},...,\cP'_{N_{p}-1}\}$
 \Ensure  Final Clusters $\cC$
  \State Cluster set $\cC = \emptyset$, Image set $\cI = \emptyset$, $i = 1$,
   \While{$i \leq N_{p}$}
  \State $\cC_{i} = \cP'_{i} \setminus \cI$
  \State $\cC = \cC \cup  \{\cC_{i}\}$
  \State $\cI = \cI \cup \cC_{i}$
  \State $i = i + 1$
 \EndWhile
\State \Return $\cC$
 \end{algorithmic}
 \label{alg:de_overlap}
\end{algorithm}

% !TEX root = ../submission.tex

\section{Experiments}
\subsection{Experimental Settings}

%% Clustering and proposal strategy

\begin{figure*}[t]
 \begin{minipage}[b]{0.7\textwidth}
  \centering\small
	\begin{tabular}{c|c|c|c|c|c}
	\hline
	Methods & \#clusters & Precision & Recall & F-score & Time \\\hline\hline
    K-Means~\cite{lloyd1982least} & 5000 & 52.52 & 70.45 & 60.18 & 13h \\
    % Mini-batch K-Means & 5000 & 45.48 & 80.98 & 58.25 & 600s\\\hline
    DBSCAN~\cite{ester1996density} & 352385 & 72.88 & 42.46 & 53.5 & \textbf{100s}\\
    % HDBSCAN & 15734 & 0.11 & 61.21 & 0.214 & 49h\\\hline
    HAC~\cite{sibson1973slink} & 117392 & 66.84 & 70.01 & 68.39 & 18h\\
    Approximate Rank Order~\cite{otto2018clustering} & 307265 & 81.1 & 7.3 & 13.34 & 250s\\
    CDP~\cite{zhan2018consensus} & 29658 & 80.19 & 70.47 & 75.01 & 350s\\\hline\hline
    \textbf{GCN-D} & 19879 & 95.72 & 76.42 & 84.99 & 2000s\\
    \textbf{GCN-D + GCN-S} & 19879 & 98.24 & 75.93 & \textbf{85.66} & 2200s\\\hline
	\end{tabular}
	\captionof{table}{
	\small Comparison on face clustering. (MS-Celeb-1M)
	}
	\label{tab:exp_cmp_cluster}
 \end{minipage}
 \hfill
 \begin{minipage}[b]{0.3\textwidth}
  \centering
	\includegraphics[width=\linewidth]{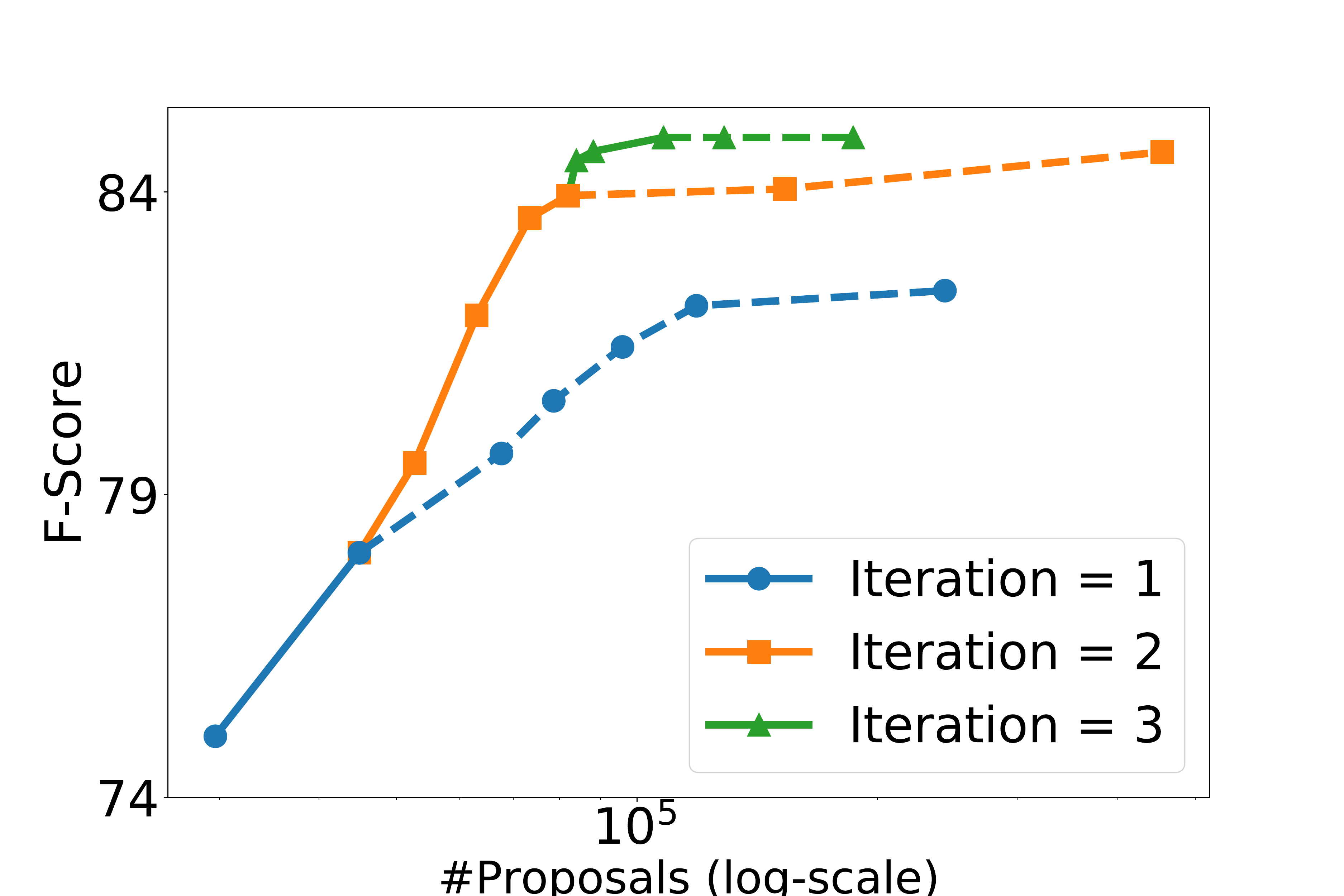}
	\captionof{figure}{\small
        Proposal strategies.
    }
	\label{fig:abl_prop_strategy}
 \end{minipage}
\end{figure*}

\noindent\textbf{Training set.}
MS-Celeb-1M~\cite{guo2016ms} is a large-scale face recognition dataset
consists of $100K$ identities, and each identity has about $100$
facial images.
As the original identity labels are obtained automatically from
webpages and thus are very noisy. We clean the labels based on the
annotations from ArcFace~\cite{deng2018arcface}, yielding a reliable subset that
contains $5.8M$ images from $86K$ classes.
The cleaned dataset is randomly split into $10$ parts with an almost
equal number of identities. Each part contains $8.6K$ identities
with around $580K$ images.
We randomly select $1$ part as labeled data and the other $9$ parts
as unlabeled data.
Youtube Face Dataset~\cite{wolf2011face} contains $3,425$ videos,
from which we extract $155,882$ frames for evaluation. Particularly,
we use $14,653$ frames with $159$ identities for training and
the other $140,629$ images with $1,436$ identities for testing.

\noindent\textbf{Testing set.} MegaFace~\cite{kemelmacher2016megaface} is the largest
public benchmark for face recognition.
It includes a probe set from FaceScrub~\cite{ng2014data} with $3,530$ images
and a gallery set containing $1M$ images.
IJB-A~\cite{klare2015pushing} is another face recognition benchmark containing
$5,712$ images from $500$ identities.

\noindent\textbf{Metrics.} We assess the performance on two tasks,
namely \emph{face clustering} and \emph{face recognition}.
Face clustering is to cluster all the images of the same identity into
a cluster, where the performance is measured by \emph{pairwise}
recall and \emph{pairwise} precision.
To consider both precision and recall, we report the widely used
\emph{F-score}, \ie, the harmonic mean of precision and recall.
Face recognition is evaluated with \emph{face identification}
benchmark in MegaFace and \emph{face verification} protocol of IJB-A.
We adopt top-$1$ identification hit rate in MegaFace, which is to
rank the top-$1$ image from the $1M$ gallery images and compute the
top-$1$ hit rate.
For IJB-A, we adopt the protocol of face verification, which is to
determine whether two given face images are from the same identity. We
use \emph{true positive rate} under the condition that the
\emph{false positive rate} is $0.001$ for evaluation.

\noindent\textbf{Implementation Details.}
We use GCN with two hidden layers in our experiments. The momentum
SGD is used with a start learning rate $0.01$.
Proposals are generated by $e_\tau \in \{0.6, 0.65, 0.7, 0.75\}$ and
$s_{max} = 300$ as in Alg. \ref{alg:super_vertex}.

\begin{table}[t]
	\centering\small
	\begin{tabular}{c|c|c|c|c|c}
	\hline
	Method & \#clusters & Precision & Recall & F-score & Time \\\hline\hline
	K-means & 1000 & 60.65 & 59.6 & 60.12 & 39min\\
	HAC & 3621 & 99.64 & 87.31 & 93.07 & 1h\\
	CDP & 3081 & 98.32 & 89.84 & 93.89 & 175s\\\hline\hline
	\textbf{Ours} & 2369 & 96.75 & 92.27 & \textbf{94.46} & 537s\\\hline
	\end{tabular}
	\caption{
	\small Comparison on face clustering. (YTF)
	}
	\label{tab:exp_cmp_cluster_ytb}
\end{table}

\subsection{Method Comparison}

\subsubsection{Face Clustering}

\vspace{-7pt}

We compare the proposed method with a series of clustering baselines.
These methods are briefly described below.

\noindent\textbf{(1) K-means~\cite{lloyd1982least},} the most commonly used clustering
algorithm. With a given number of clusters $k$, K-means minimizes
the total intra-cluster variance.
%
%Mini-batch K-means, a variant of K-means, achieves compatible result
%with K-means but a significant less of time.

%
\noindent\textbf{(2) DBSCAN~\cite{ester1996density},} a density-based clustering
algorithm. It extracts clusters according to a designed density criterion and leaves the sparse background as noises.
%HDBSCAN~\misscite adapts DBSCAN to allow varying density clusters.

%
\noindent\textbf{(3) HAC~\cite{sibson1973slink},}
hierarchical agglomerative clustering is a bottom-up approach to
iteratively merge close clusters based on some criteria.

\if 0
%
\noindent\textbf{Spectral Clustering~\misscite,}  demands a symmetric
	$N\times N$ matrix for eigenvalue decomposition. It suffers from
	the same memory problem as HAC in our scenario.
\fi

%
\noindent\textbf{(4) Approximate Rank Order~\cite{otto2018clustering},}
develops an algorithm as a form of HAC. It only performs one iteration
of clustering with a modified distance measure.

\noindent\textbf{(5) CDP~\cite{zhan2018consensus},} a recent work that proposes a
graph-based clustering algorithm. It better exploits the pairwise
relationship in a bottom-up manner.

\noindent\textbf{(6) GCN-D,} the first module of the proposed method.
It applies a GCN to learn cluster pattern in a supervised way.

\noindent\textbf{(7) GCN-D + GCN-S,} the two-stage version of the
proposed method. GCN-S is introduced to refine the output of GCN-D,
which detects and discards noises inside clusters.

\begin{figure*}[h]
	\centering
    \begin{subfigure}[t]{0.3\linewidth}
        \centering
        \includegraphics[width=\linewidth]{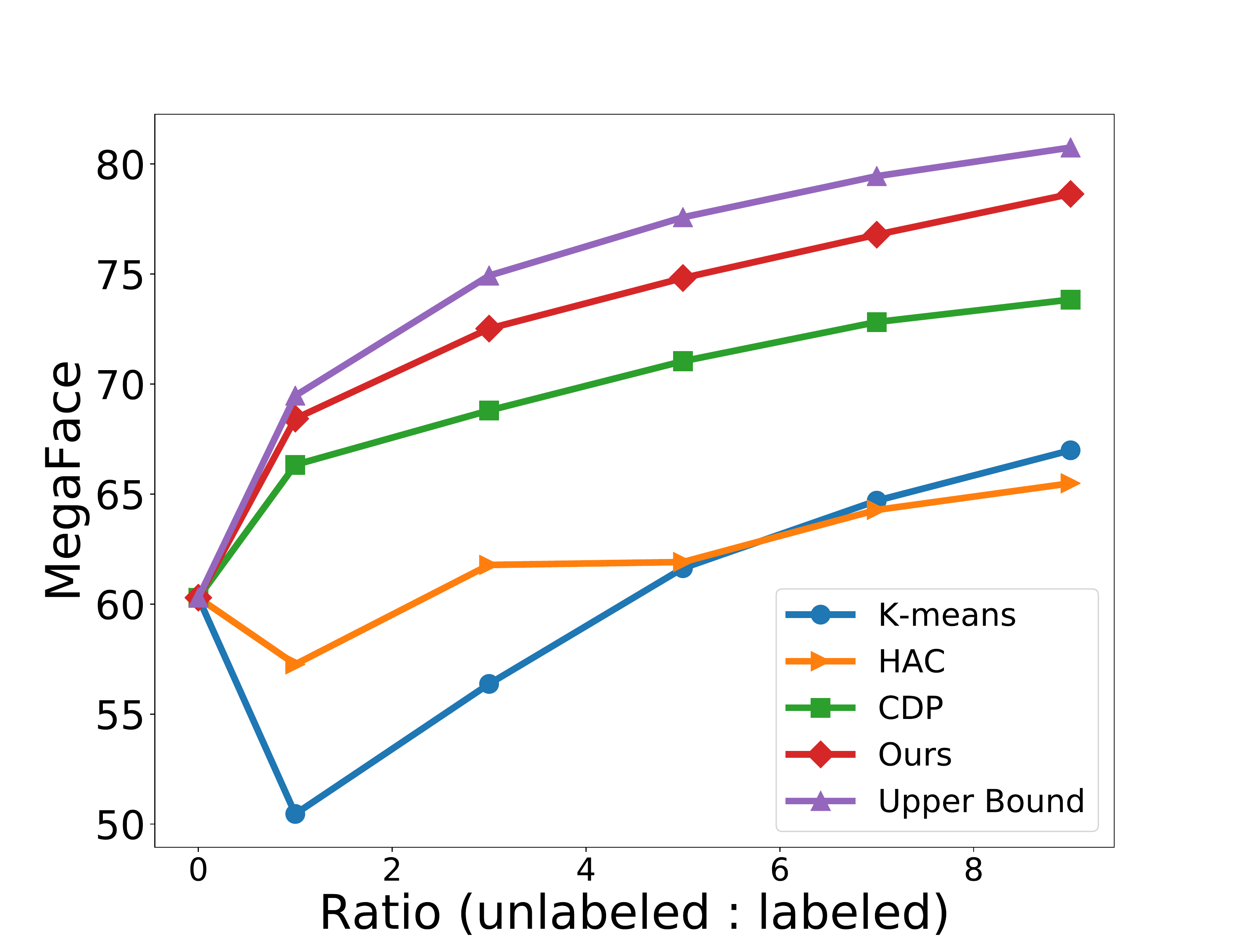}
        \caption{MegaFace top-1 Identification@1M}
    \end{subfigure}%
    ~
    \begin{subfigure}[t]{0.3\linewidth}
        \centering
        \includegraphics[width=\linewidth]{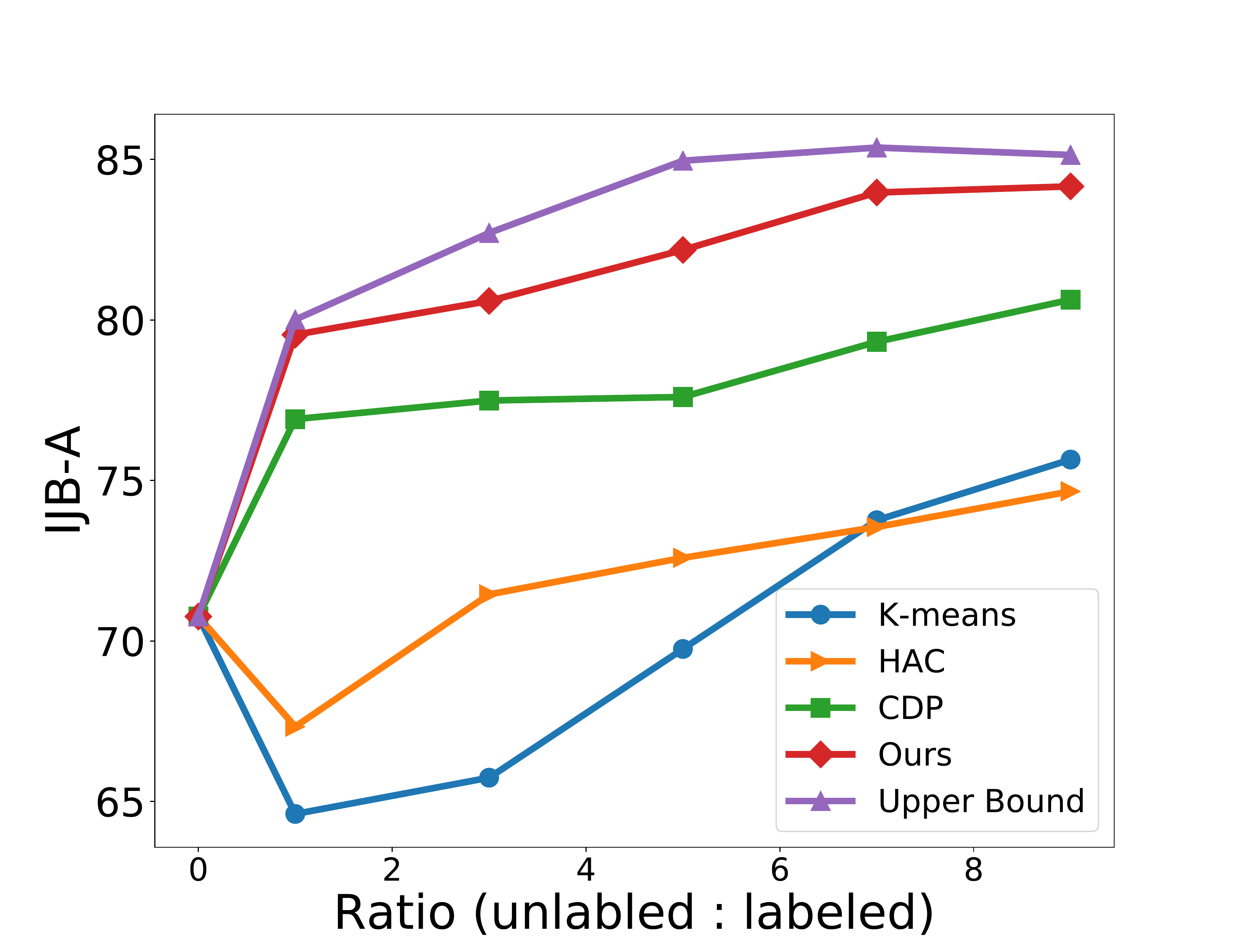}
        \caption{IJB-A TPR@FPR=0.001}
    \end{subfigure}
    \caption{\small
    Comparison on face recognition. The performance is measured by
        MegaFace Identification and IJB-A Verification.
    The leftmost point is the performance obtained when only labeled data are used;
	the rightmost point is MegaFace performance obtained when both labeled and unlabeled data are used.
    }

	\label{fig:comparison}
\end{figure*}

\vspace{-11pt}
\paragraph{Results}
To control the experimental time, we randomly select one part of the
data for evaluation, containing $580K$ images of $8,573$ identities.
Tab.~\ref{tab:exp_cmp_cluster} compares the performance of different
methods on this set. The clustering performance is evaluated by both F-score and the time cost. We also report the number of clusters,
pairwise precision and pairwise recall for better understanding the
advantages and disadvantages of each method.

\begin{figure}[t]
	\centering
	\includegraphics[width=\linewidth]{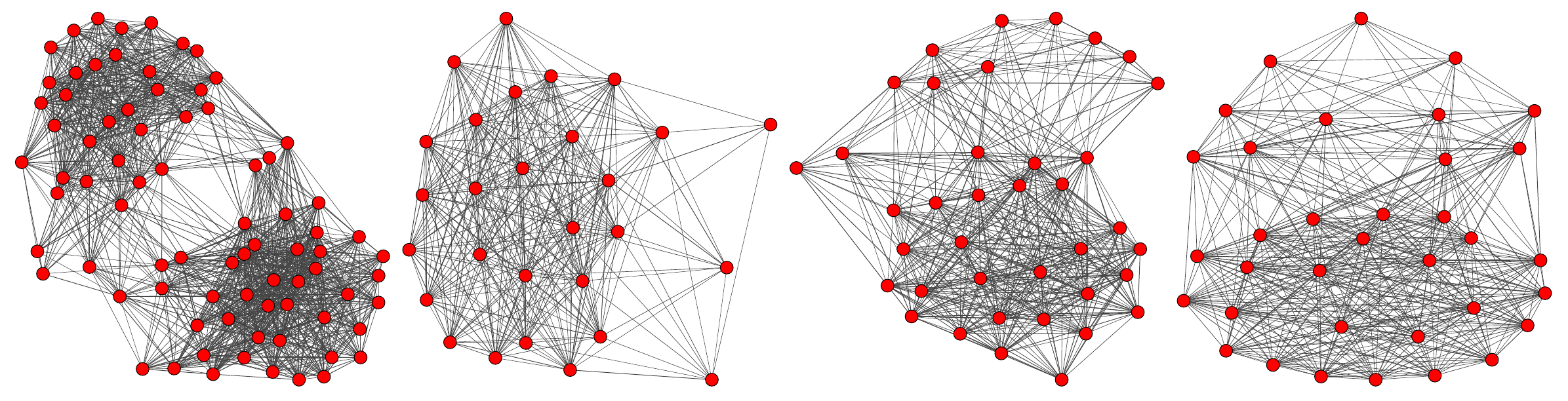}
	\caption{\small
        The figure shows 4 clusters pinpointed by our approach.
        All vertices in each cluster belong to the same class
        according to ground-truth annotation.
        The distance between vertices is inversely proportional
        to the similarity between vertices.
        It shows that our method can handle clusters with complex
        intra-structure, \eg clusters with two sub-graphs inside,
        clusters with both dense and sparse connections.
    }
	\label{fig:vis_props}
\end{figure}

The results show:
% K-means
(1) For K-means, the performance is influenced greatly by
the number of clusters $k$. We vary $k$ in a range of numbers and
report the result with high F-score.
%Our method still outperforms K-means without the prior of the number of clusters.
% DBSCAN
(2) DBSCAN reaches a high precision but suffers from the
low recall. It may fail to deal with large density differences in large-scale face clustering.
% HAC
(3) HAC gives more robust results than previous methods.
Note that the standard algorithm consumes $O(N^2)$ memory, which goes
beyond the memory capacity when $N$ is as large as $580K$. We use an adapted hierarchical
clustering~\cite{mullner2013fastcluster} for comparison, which requires only $O(Nd)$ memory.
% ARO
(4) Approximate Rank Order is very efficient due to its
one iteration design, but the performance is
inferior to other methods in our setting.
% CDP
(5) As a recent work designed to exploit unlabeled data for face
recognition, CDP achieves a good balance of precision and recall.
For a fair comparison, we compare with the single model version of CDP.
Note that the idea of CDP and our approach are
complementary, which can be combined to further improve the performance.
% Ours
(6) Our method applies GCN to learn cluster patterns.
It improves the precision and recall simultaneously.
Tab.~\ref{tab:exp_cmp_cluster_ytb} demonstrates that our method is robust
and can be applied to datasets with different distributions.
Since the GCN is trained using multi-scale cluster proposals,
it may better capture the properties of the desired clusters.
As shown in Fig.~\ref{fig:vis_props}, our method is capable of
pinpointing some clusters with complex structure.
(7) The GCN-S module further refines the cluster proposals from the
first stage. It improves the precision by sacrificing a little
recall, resulting in the overall performance gain.
%
%Fig.~\ref{fig:vis_clusters} gives a qualitative view of our method. It illustrates that our approach has high precision in grouping images even when there is large intra-class variance, \eg, pose,
%age, facial expression. Besides, our method also excels at pinpointing wrongly annotated images (first two rows in
%Fig.~\ref{fig:vis_clusters}) and excluding some low-quality images,
%\eg heavy makeup, heavy occlusion (last two rows in
%Fig.~\ref{fig:vis_clusters}), which may affect the training of face recognition.

\vspace{-11pt}
\paragraph{Runtime Analysis}
The whole procedure of our method takes about $2200s$, where generating $150K$ proposals
takes up to $1000s$ on a CPU and the inference of GCN-D and GCN-S
takes $1000s$ and $200s$ respectively on a GPU with the batch size of $32$.
To compare the runtime fairly, we also test all our modules on CPU.
Our method takes $3700s$ in total on CPU, which is still faster than
most methods we compared.
The speed gain of using GPU is not very significant in this work,
as the main computing cost is on GCN.
Since GCN relies on sparse matrix multiplication, it cannot make full use of GPU parallelism.
%Since only a small amount of clusters are predicted to have noise in
%our desired range, GCN-S only increases a bit more computation cost.
%
The runtime of our method grows linearly with the number of
unlabeled data and the process can be further accelerated by
increasing batch size or parallelizing with more GPUs.

\vspace{-11pt}
\subsubsection{Face Recognition}

With the trained clustering model, we apply it to unlabeled data
to obtain pseudo labels. We investigate how the unlabeled data with
pseudo labels enhance the performance of face recognition.
%In particular, we use 1 part of data to train an initial face
%recognition model as well as the proposed clustering model, and then
%improve the face model adding various amount (1, 3, 5, 7, 9 parts) of
%unlabeled data with pseudo labels.
Particularly, we follow the following steps to train face recognition models:
(1) train the initial recognition model with labeled data in a supervised way;
(2) train the clustering model on the labeled set, using the feature
representation derived from the initial model;
(3) apply the clustering model to group unlabeled data with various
amounts (1, 3, 5, 7, 9 parts), and thus attach to them \emph{``pseudo-labels''}; and
(4) train the final recognition model using the whole dataset, with
both original labeled data and the others with assigned pseudo-labels.
The model trained only on the 1 part labeled data is regarded as the lower
bound, while the model supervised by all the parts with ground-truth labels serves as the
upper bound in our problem.
For all clustering methods, each unlabeled image belongs to an
unique cluster after clustering. We assign a pseudo label to
each image as its cluster id.
%Then we re-train the model from scratch
%using both labeled data and unlabeled data with assigned labels, in
%a multi-task manner. % Do we need to explain multitask?

%
Fig.~\ref{fig:comparison} indicates that performance of face
clustering is crucial for improving face recognition.
For K-means and HAC, although the recall is good, the low precision
indicates noisy predicted clusters.
When the ratio of unlabeled and labeled data is small, the noisy
clusters severely impair face recognition training.
As the ratio of unlabeled and labeled data increases,
the gain brought by the increase of unlabeled data alleviates
the influence of noise.
However, the overall improvement is limited.
Both CDP and our approach benefit from the increase of the
unlabeled data.
Owing to the performance gain in clustering, our approach outperforms
CDP consistently and improve the performance of face recognition model on
MegaFace from $60.29$ to $78.64$, which is close to the fully
supervised upper bound ($80.75$).

\subsection{Ablation Study}

We randomly select one part of the unlabeled data, containing $580K$
images of $8,573$ identities, to study some important design
choices in our framework.

\vspace{-11pt}
\subsubsection{Proposal Strategies}
\label{sec:abl_props}

Cluster proposals generation is the fundamental module in our framework.
With a fixed $K=80$ and different $I$, $e_\tau$ and $s_{max}$,
we generate a large number of proposals with multiple scales.
Generally, a larger number of proposals result in a better
clustering performance. There is a trade-off between performance and
computational cost in choosing the proper number of proposals.
As illustrated in Fig.~\ref{fig:abl_prop_strategy},
each point represents the F-score under certain number of proposals.
Different colors imply different iteration steps.
(1) When $I=1$, only the super-vertices generated by
Alg.~\ref{alg:super_vertex} will be used. 
By choosing different $e_\tau$, more proposals are obtained to
increase the F-score.
The performance gradually saturates as the number increases beyond $100K$. 
(2) When $I=2$, different combinations of super-vertices are
added to the proposals. Recall that it leverages the similarity between
super-vertices, thus it enlarges the receptive
field of the proposals effectively. With a small number of
proposals added, it boosts the F-score by $5\%$.
(3) When $I=3$, it further merges similar proposals from previous
stages to create proposals with larger scales, which continues to
contribute the performance gain. However, with the increasing
proposal scales, more noises will be introduced to the proposals,
hence the performance gain saturates.

\begin{table}[t]
	\centering\small
	\begin{tabular}{c|c|c|c|c}
	\hline
	Method & Channels & Pooling & \makecell{Vertex \\ Feature} & F-score \\\hline\hline
    a & 128, 32   & mean & \checkmark & 76.97 \\\hline
    b & 128, 32   & sum & \checkmark & 53.75 \\\hline
    c & 128, 32   & max & \checkmark & 83.61 \\\hline
    d & 128, 32   & max & \texttimes & 73.06 \\\hline
    e & 256, 64   & max & \checkmark & 84.16 \\\hline
    f & 256, 128, 64   & max & \checkmark & 77.95 \\\hline
	\end{tabular}
	\caption{
	\small Design choice of GCN-D
	}
	\label{tab:exp_abl_gcn_d}
\end{table}

\vspace{-12pt}
\subsubsection{Design choice of GCN-D}

Although the training of GCNs does not require any fancy techniques,
there are some important design choices.
As Tabs.~\ref{tab:exp_abl_gcn_d}a, ~\ref{tab:exp_abl_gcn_d}b and ~\ref{tab:exp_abl_gcn_d}c indicate, the pooling method has large
influence on the F-score.
Both mean pooling and sum pooling impair the clustering
results compared with max pooling. For sum pooling, it is sensitive to
the number of vertices, which tends to produce large proposals.
Large proposals result in a high recall($80.55$) but low precision
($40.33$), ending up with a low F-score. On the other hand, mean
pooling better describes the graph structures, but
may suffer from the outliers in the proposal.
Besides the pooling methods, Tabs.~\ref{tab:exp_abl_gcn_d}c and ~\ref{tab:exp_abl_gcn_d}d
show that lacking vertex feature will significantly
reduce the GCNs' prediction accuracy.
It demonstrates the necessity of
leveraging both vertex feature and graph structure during GCN training.
In addition, as shown in Tabs.~\ref{tab:exp_abl_gcn_d}c, ~\ref{tab:exp_abl_gcn_d}e and ~\ref{tab:exp_abl_gcn_d}f,
widening the channels of GCNs can increase its
expression power but the deeper network may drive the hidden feature of
vertices to be similar, resulting in an effect like mean pooling.

\begin{figure}[t]
 \begin{minipage}[b]{0.48\linewidth}
  \centering
  \includegraphics[width=\linewidth]{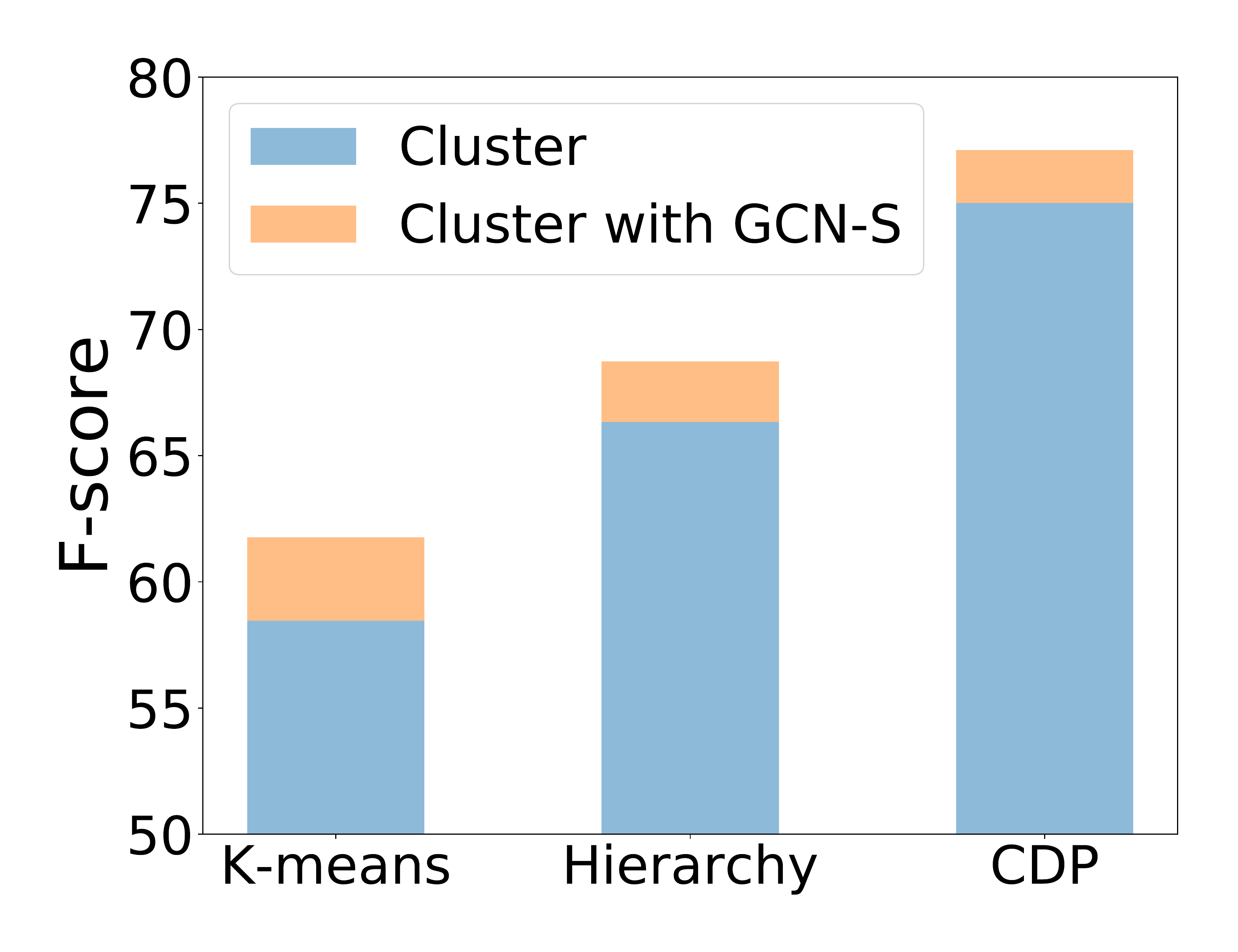}
  \caption{GCN-S}
  \label{fig:abl_gcn_s}
 \end{minipage}
 \begin{minipage}[b]{0.48\linewidth}
  \centering
  \includegraphics[width=\linewidth]{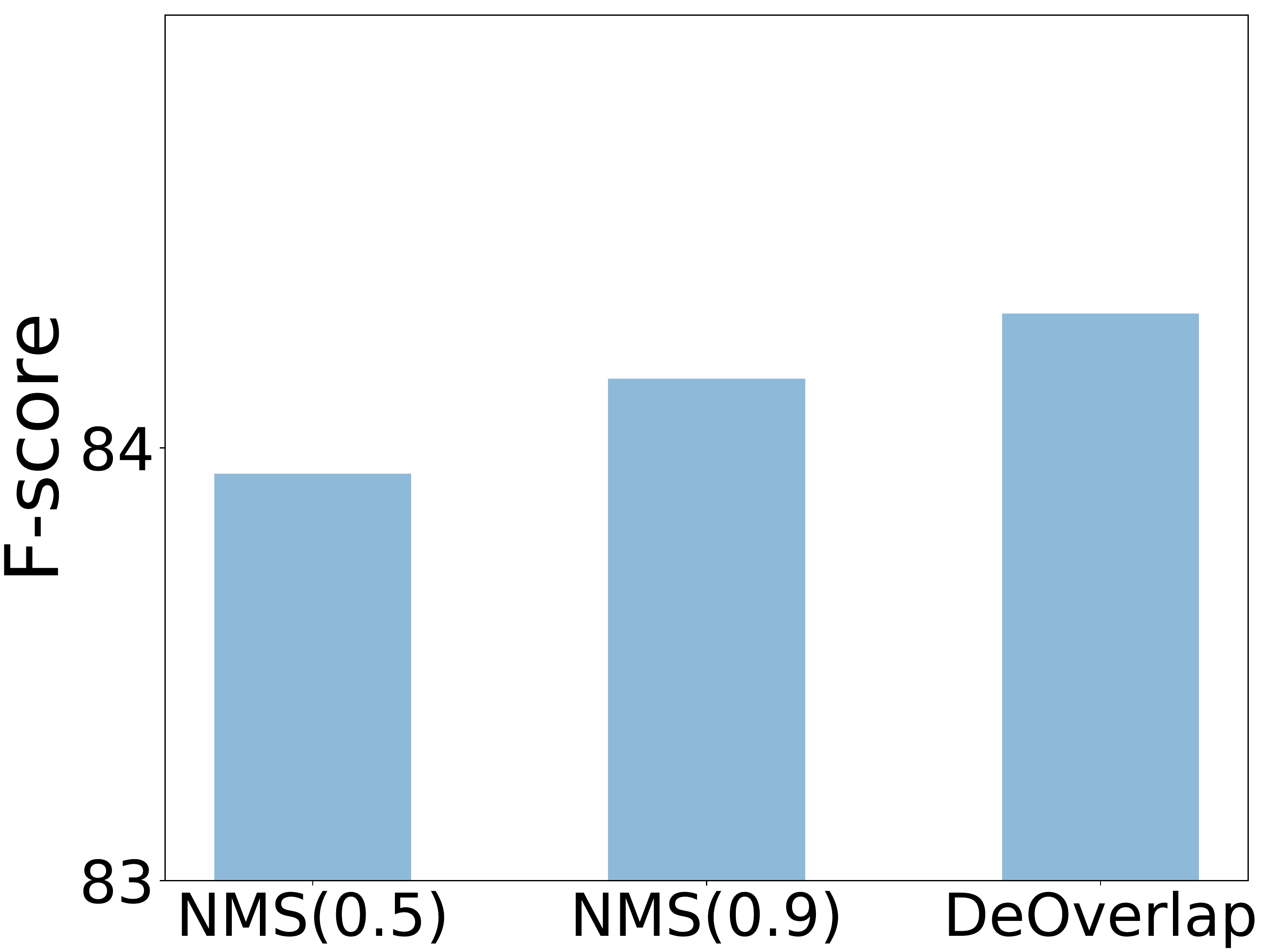}
  \caption{Post-processing}
  \label{fig:exp_abl_post}
 \end{minipage}
\end{figure}

\vspace{-13pt}
\subsubsection{GCN-S}
In our framework, GCN-S is used as a de-nosing module after GCN-D.
However, it can act as an independent module to combine with
previous methods.
Given the clustering results of
K-means, HAC and CDP, we regard them as the cluster proposals and
feed them into the GCN-S.
As Fig.~\ref{fig:abl_gcn_s} shows, GCN-S can improve their clustering
performances by discarding the outliers inside clusters,
obtaining a performance gain around $2\%-5\%$ for various methods.

\vspace{-15pt}
\subsubsection{Post-process strategies}
NMS is a widely used post-processing technique in object detection,
which can be an alternative choice of de-overlapping.
With a different threshold of IoU, it keeps the proposal with
highest predicted IoU while suppressing other overlapped proposals.
The computational complexity of NMS is $O(N^2)$.
Compared with NMS, de-overlapping does not suppress other proposals
and thus retains more samples, which increases the clustering recall.
As shown in Fig.~\ref{fig:exp_abl_post}, de-overlapping achieves better clustering
performance and can be computed in linear time.

%\subsection{Further Analysis}

%\subsection{Visualization}

%\begin{figure}[t]
%	\centering
%	\includegraphics[width=\linewidth]{./figs/vis_clusters}
%	\caption{\small
%        Visualization of clustering results on unlabeled data.
%        Each row shows a cluster predicted by our results.
%        Images with red box indicates they belong to the same class
%        in ground-truth annotation but discarded by our method.
%        }
%	\label{fig:vis_clusters}
%\end{figure}

% !TEX root = ../submission.tex

\section{Conclusions}

This paper proposes a novel supervised face clustering framework based on
graph convolution network.
Particularly, we formulate clustering as a detection and segmentation
paradigm on an affinity graph.
The proposed method outperforms previous methods on face clustering
by a large margin, which consequently boosts the face recognition
performance close to the supervised result.
Extensive analysis further
demonstrate the effectiveness of our framework.
%
%Specifically, the GCN-S can be easily plugged into previous clustering
%methods for further improvement.

\vspace{7pt}
\noindent \textbf{Acknowledgement} This work is partially supported by the Collaborative Research grant from SenseTime Group (CUHK Agreement No. TS1610626 \& No. TS1712093), the Early Career Scheme (ECS) of Hong Kong (No. 24204215), the General Research Fund (GRF) of Hong Kong (No. 14236516, No. 14203518 \& No. 14241716), and Singapore MOE AcRF Tier 1 (M4012082.020).

{\small
\bibliographystyle{ieee_fullname}
\bibliography{egbib}
}

\end{document}